\documentclass[twoside,leqno,twocolumn]{article}

\usepackage[letterpaper]{geometry}

\usepackage{ltexpprt}

\usepackage{sidecap}

\usepackage{cite}
\usepackage{amsmath,amssymb,amsfonts}
\usepackage{algorithmic}
\usepackage{graphicx}
\usepackage{textcomp}

\usepackage{xcolor}

\usepackage{comment}
\usepackage{amssymb}
\usepackage{bbm}
\usepackage{subfig}
\usepackage{booktabs}
\usepackage{multirow}
\usepackage[export]{adjustbox}
\usepackage{todonotes}
\usepackage{amsmath} 
\usepackage{algorithm}
\usepackage[algo2e,linesnumbered,ruled,vlined]{algorithm2e}

\begin{document}

\title{\Large MLCTR: A Fast Scalable Coupled Tensor Completion Based on Multi-Layer Non-Linear Matrix Factorization}
\author{Ajim Uddin\thanks{New Jersey Institute of Technology, Newark, NJ, USA. \{au76, dz239, xinyuan.tao, dtyu\}.njit.edu}
\and Dan Zhou\footnotemark[1]
\and Xinyuan Tao\footnotemark[1]
\and Chia-Ching Chou\thanks{Central Michigan University, Mount Pleasant, MI, USA. chou1c@cmich.edu}
\and Dantong Yu\footnotemark[1]
}

\date{}

\maketitle


\fancyfoot[R]{\scriptsize{Copyright \textcopyright\ 20XX by SIAM\\
Unauthorized reproduction of this article is prohibited}}





\begin{abstract} \small\baselineskip=9pt 
Firms earning prediction plays a vital role in investment decisions, dividends expectation, and share price. It often involves multiple tensor-compatible datasets with non-linear multi-way relationships, spatiotemporal structures, and different levels of sparsity. 
Current non-linear tensor completion algorithms tend to learn noisy embedding and incur overfitting. 
This paper focuses on the embedding learning aspect of the tensor completion problem and proposes a new multi-layer neural network architecture for tensor factorization and completion (MLCTR).  The network architecture entails multiple advantages: a series of low-rank matrix factorizations (MF) building blocks to minimize overfitting, interleaved transfer functions in each layer for non-linearity, and by-pass connections to reduce the gradient diminishing problem and increase the depths of neural networks.  Furthermore, the model employs  Stochastic Gradient Descent (SGD) based optimization for fast convergence in training. Our algorithm is highly efficient for imputing missing values in the EPS data. Experiments confirm that our strategy of incorporating non-linearity in factor matrices demonstrates impressive performance in embedding learning and end-to-end tensor models, and outperforms approaches with non-linearity in the phase of reconstructing tensors from factor matrices.\end{abstract} 


\section{Introduction}


\begin{figure}
\begin{centering}
  \includegraphics[width=.70\linewidth]{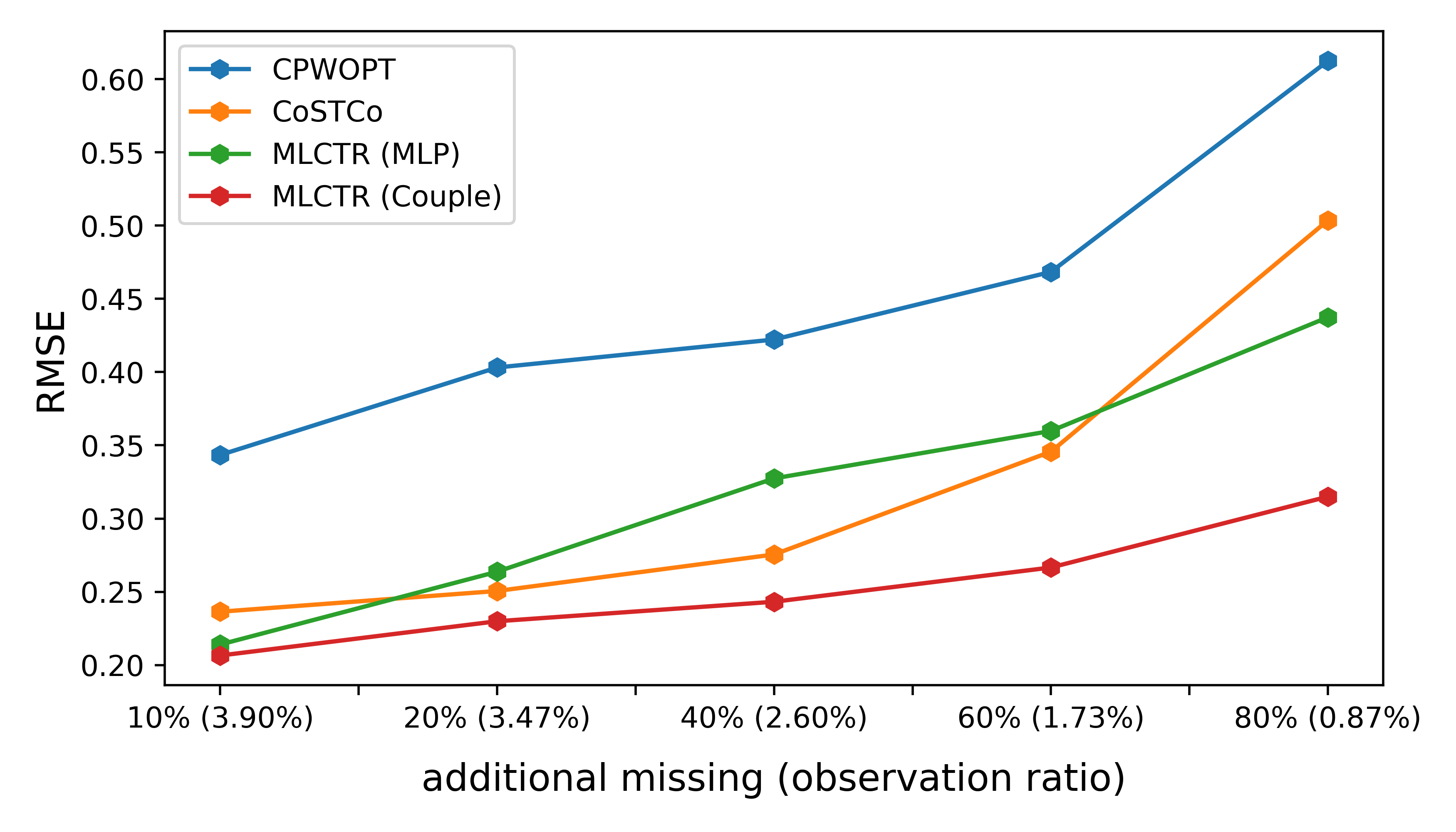}
  \vspace{-0.1in}
    \caption{Tensor completion on the Earning Per Share (EPS) data. The accuracy of tensor completion with low-rank decomposition severely degenerates with increasing tensor sparsity. In contrast, our algorithm with the coupled tensor factorization maintains accuracy even with 99.13\% of missing values.}
  \label{fig:tens_com}
  \vspace{-0.2in}
 \end{centering}
\end{figure}

\begin{figure*}
\vspace{-1.3cm}
  \centering
  \subfloat[CP]{\includegraphics[trim=0 1cm 0 1.5cm, clip=true, width=0.31\textwidth]{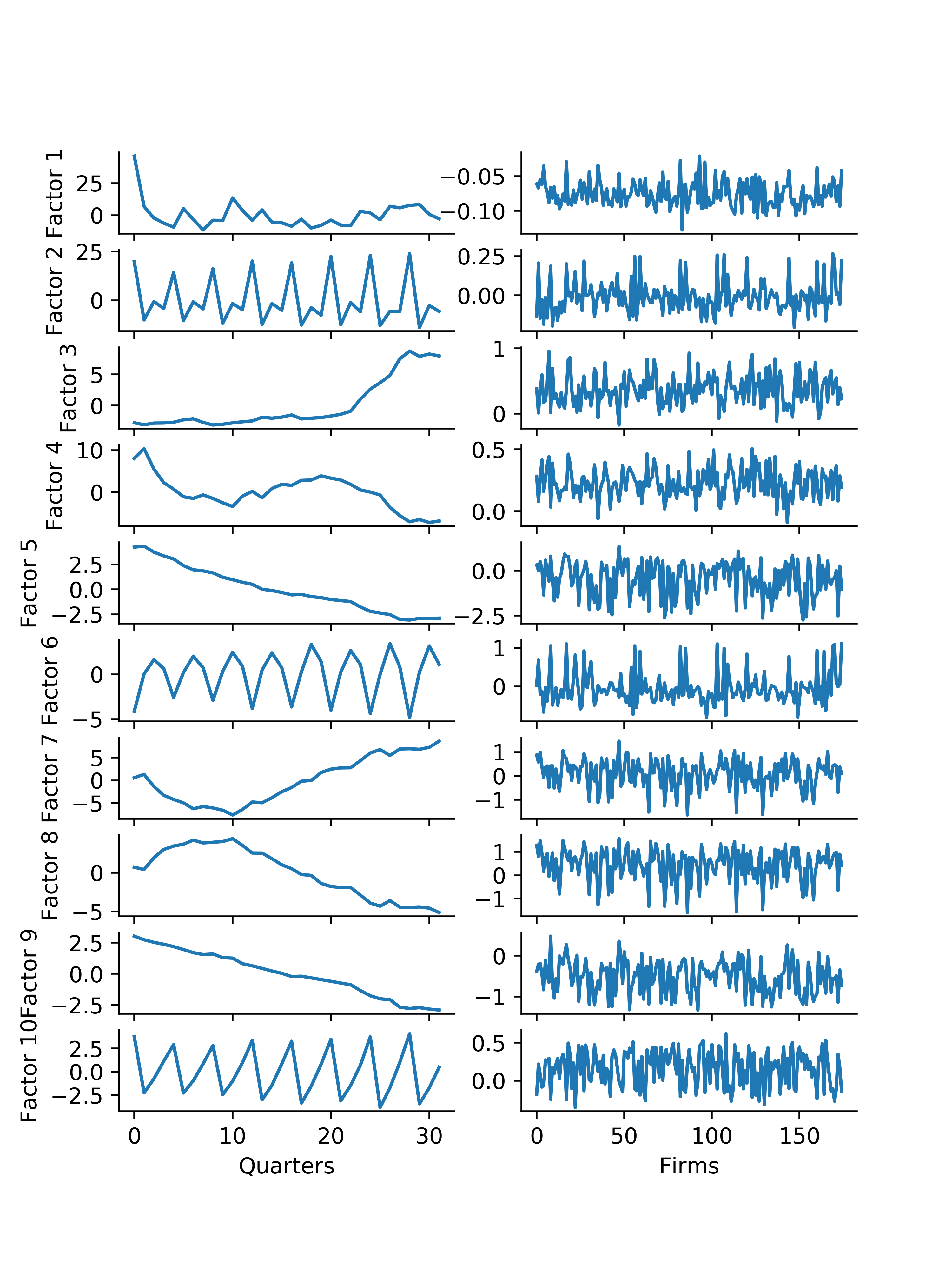}\label{fig:CP_f}}
  \subfloat[CoSTCo]{\includegraphics[trim=0 1cm 0 1.5cm, clip=true, width=0.31\textwidth]{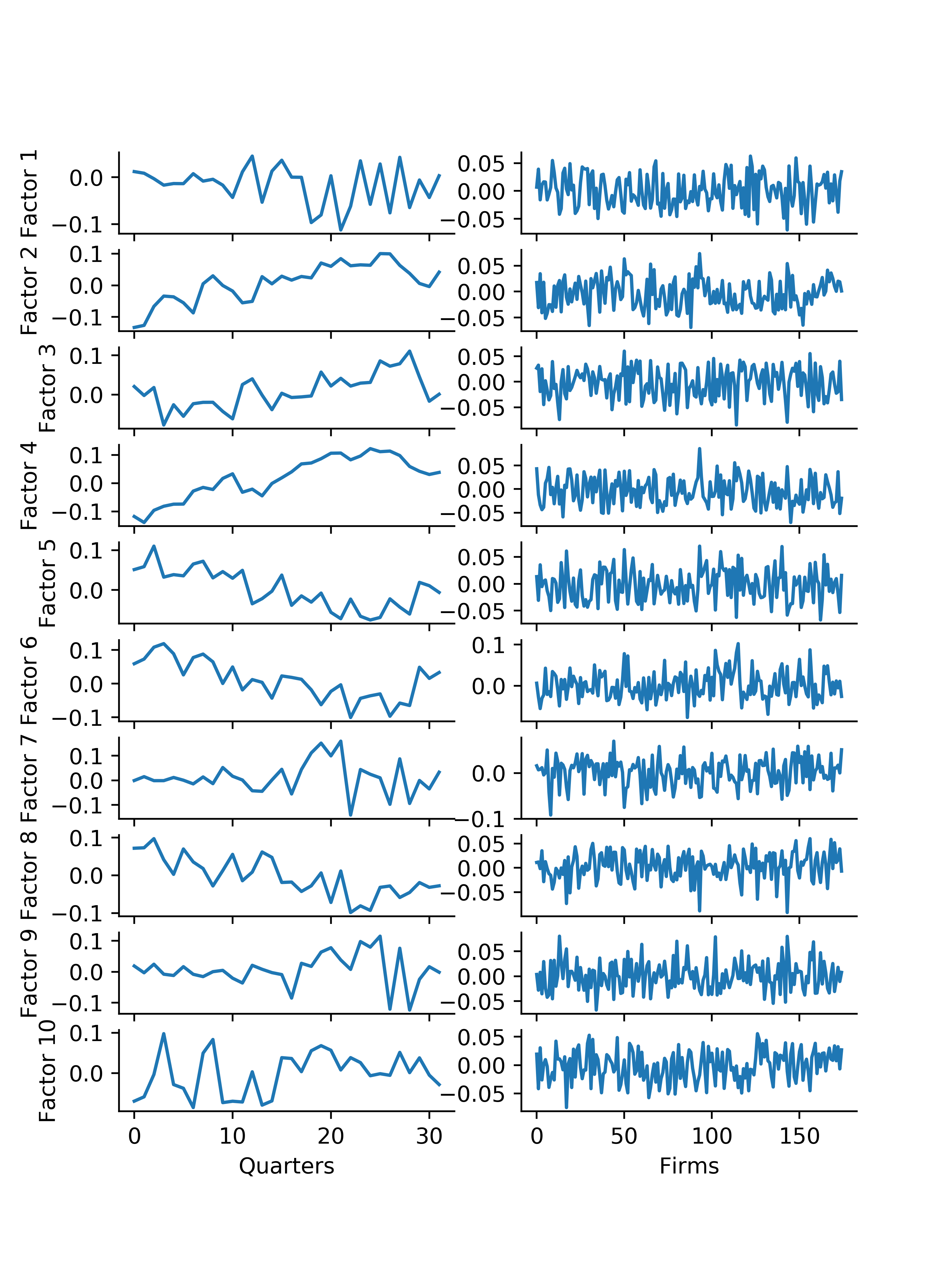}\label{fig:COS_F}}
    \subfloat[MLCTR]{\includegraphics[trim=0 1cm 0 1.5cm, clip=true, width=0.31\textwidth]{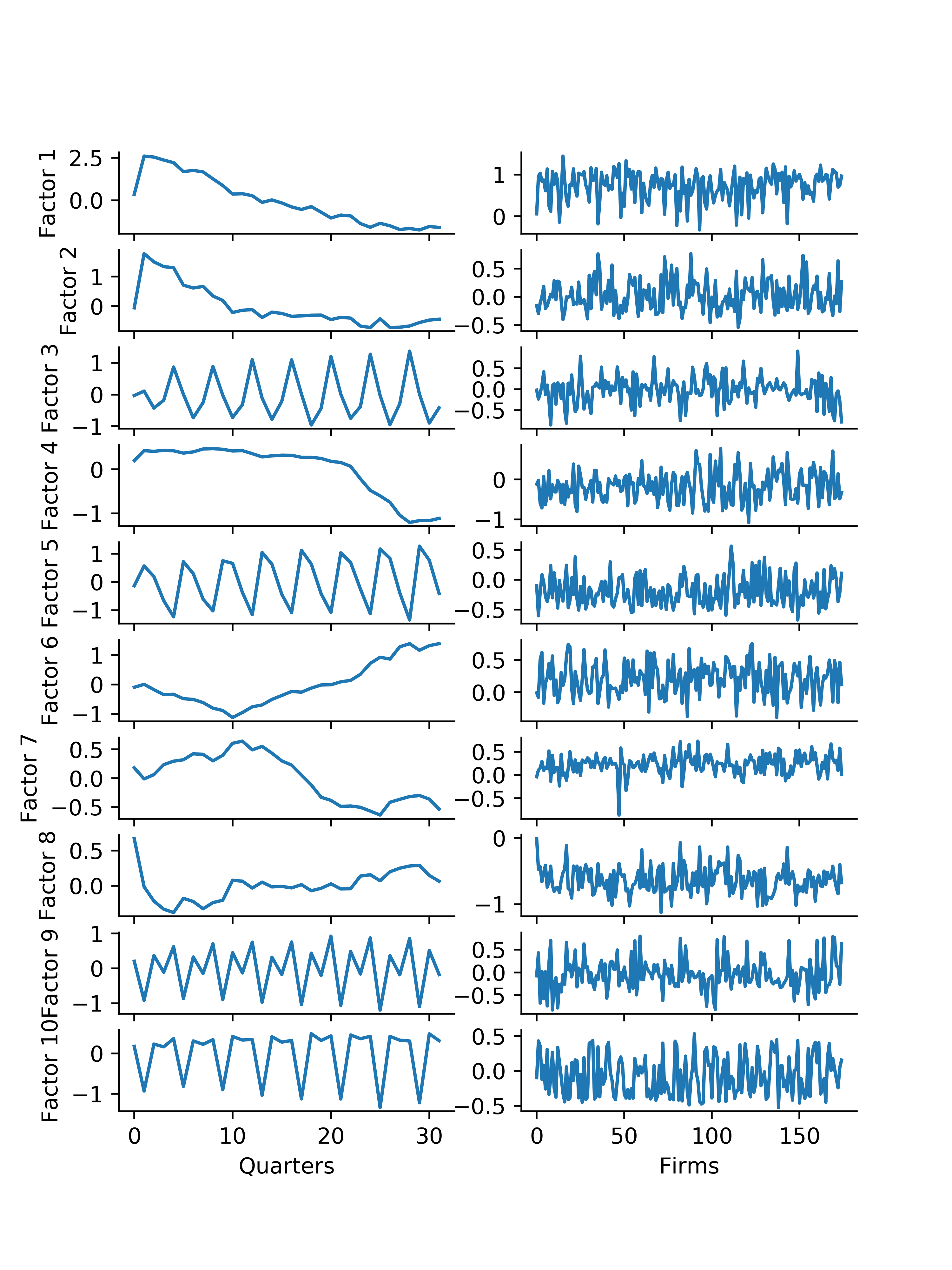}\label{fig:MLCTR}}
   \caption{The quarters and firms embedding matrix learned from (a) CP based factorization, (b) CoSTCo, and (c) MLCTR. The factors in both CP and MLCTR are much more clear and concise than CoSTCo. Especially, the time embedding factors clearly demonstrate patterns in the high frequency band and low frequency band and  captures   the yearly business cycles (8 peaks and trough for 32 quarters) and US economic recovery from 2011-2012 onward (MLCTR 6th and 7th factors). In CP and MLCTR factorization, almost all information is forced to pass through the embedding matrices, whereas in CoSTCo the CNN of the later part also captures a significant portion of tensor information, resulting in less informative embedding matrices.
   }
  \label{fig:embedding}
    \vspace{-0.1in}
\end{figure*}

Tensor completion algorithms are mostly based on two representative low-rank tensor factorization models:  CANDECOMP/PARAFAC (CP) \cite{harshman1970foundations} and Tucker \cite{tucker1966some}.  These approaches attempt to identify low-rank factor matrices using the observed entries and then reconstruct the target tensor based on these factors matrices. The low-rank tensor completion essentially boils down to a two-step problem with two objectives: representation learning (Factorization) and subsequent multi-way relationship prediction (Reconstruction).  Over the years, a number of low-rank tensor completion methods have been proposed \cite{acar2011scalable, gandy2011tensor,  song2017multi, liu2018neuralcp, wu2019neural, liu2019}. 
These existing algorithms suffer two mutually exclusive problems in achieving the two objectives. First, linear low-rank algorithms  \cite{acar2011scalable, gandy2011tensor,  song2017multi}) attain low-rank embedding matrices by Singular Value Decomposition (SVD), but fail to capture the multi-way (non-linear) relationships that are common in real-world tensor applications. The lack of multi-way relationship modeling results in suboptimal performance in tensor completion or downstream prediction \cite{fang2015personalized, he2014dusk, zhe2016distributed}. 

Second, algorithms with nonlinear reconstruction (e.g.,\cite{liu2018neuralcp, wu2019neural, liu2019}) focus on nonlinear relationship learning among factors and use ``kernel tricks'' or neural network layers to represent the embedding factors and multi-way relationships.  When multi-way relationship learning becomes dominant, it ignores the structure of input signals, makes no constraints on factor matrices,  and attempts to encode all information signal into the relationships.  The lack of data structure in embedding matrices will lead to low-quality embeddings that are prone to noise, variance, and overfitting.    

In this paper, we design a  multi-layer matrix factorization neural networks for coupled tensor reconstruction (MLCTR).  MLCTR learns the distributed representations effectively for prediction and multi-way relationship tasks.  Unlike existing nonlinear tensor completion methods, it avoids the difficult trade-off between the two tensor completion objectives and introduces non-linearity in the embedding learning step.  It explicitly employs Multi-Layer matrix factorization for factor matrices and uses nonlinear transfer functions in each layer, thereby learning the highly complex structures and relationship among the hidden variables. Besides, to avoid the vanishing gradient in the deep architecture, we use by-pass connection following \cite{he2016deep}. The resulted architecture has less reconstruction error (Fig. \ref{fig:tens_com}) and generates high-quality embedding matrix (Fig. \ref{fig:embedding}).

Figure \ref{fig:tens_com} also shows the sparsity problem with tensor completion: as the number of missing observations increases, the accuracy decreases significantly. With our proposed model, we can easily mitigate this problem by augmenting sparse data sets with auxiliary data. Literature suggest auxiliary information from a secondary dataset significantly improves tensor completion accuracy\cite{narita2012tensor, kim2017discriminative, acar2011all, bahargam2018constrained}. We take advantage of data structures among multiple tensors, apply tensor integration mechanism as appropriate to reduce the associated computation cost, and scale-up MLCTR to factorize two or more coupled sparse tensors simultaneously.

Our coupled tensor completion algorithm uses a modified objective function for element-wise reconstruction and SGD optimization. To confirm the MLCTR algorithm's superiority, we evaluate it on finance datasets and three other commonly used public data sets, including climate and point of interest (POI) data. The experiment results reveal the consistency and reliability of our model.  The main contributions of our paper are as follows: 

\begin{itemize}
\item 
We develop a novel nonlinear coupled tensor completion model based on multi-layer matrix factorization, nonlinear-deep neural networks, and bypass connections to efficiently learn both embedding matrix and nonlinear interaction between the embedded vectors. 

\item The learned embeddings encode latent data structures and patterns and provide high-quality distributed representation for downstream machine learning tasks.   

\item We propose the first-ever SGD based nonlinear coupled tensor completion algorithm that is fast and scalable. 

\item We introduce the by-pass connection to mitigate the gradient diminishing problem in networks with great depths.



\end{itemize}


\section{Related Work}
\subsection{Tensor Completion}
In recent years, a number of low-rank tensor completion algorithms  \cite{gandy2011tensor, liu2012tensor, song2017multi, acar2011scalable} are developed based on classical CP \cite{harshman1970foundations} and Tucker factorization \cite{tucker1966some}. The low-rank approach is not always precise and often fails to capture frequent nonlinear interactions in real-world applications.
To capture real-world nonlinear relationships, in \cite{liu2018neuralcp, wu2019neural}, authors replace multi-linear operation with multi-layer perceptions (MLP) and in \cite{liu2019} authors proposed a convolution neural network-based architecture. Nevertheless, these works try to learn the low-rank representation of single tensor and do not consider any auxiliary information to improve the factor matrices. In \cite{narita2012tensor, kim2017discriminative}, authors introduce regularization from auxiliary data and demonstrate performance improvement. In recent years, coupled matrix-tensor factorization (CMTF) also gains broad interests \cite{acar2011all, bahargam2018constrained}. CMTF factorizes a higher-order tensor with a related matrix in a coupled fashion. Unlike CMTF, our approach is a coupled tensor factorization for sparse data where both data sets are higher-order tensors. There are several coupled tensor factorization approaches available \cite{khan2016bayesian, genicot2016coupled, wu2019improved}; however, these models are not designed for sparse data and require a full observation in both tensors. Compared to this, our MLCTR relaxes the constraints of complete observations and captures non-linearity in both target tensors. 



\subsection{EPS Forecast for Estimating Firm's Earnings}
Expected EPS conveys the vital information about firms' future cash flows and is one of the critical inputs for security pricing \cite{lee2017uncovering}. The current industry benchmark averages across all available analysts' forecasts for each firm at each quarter. However, studies suggest that this straightforward average forecast has several drawbacks: it may contain systematic bias \cite{ramnath2008financial, bradshaw2012re} and fail to incorporate additional information from markets, firm characteristics, analysts features, and crowd-sourcing \cite{bradley2017before, ball2018automated}. To address these problems, the authors in \cite{bradley2017before} assign different weights among analysts based on their past performance.  More efforts of \cite{corredor2019role, bradshaw2012re} show that the model combining accounting characteristics with analysts' EPS forecast generates better earning predictions than the widely adopted time-series models do. Our work entails a novel data mining approach MLCTR to explore a new avenue of analyzing financial data, especially with tensor representation and missing value imputation.

\section{Methodology}
Tensor factorization can be formulated as a two-step paradigm: embedding learning and subsequent relationship modeling. In contrast to the majority of tensor algorithms that focus on non-linear relationship modeling in the second step, MLCTR considers the first embedding learning step and explicitly guides networks to learn representative vectors for each entity. The MLCTR learns the embedding matrix and multi-way interaction among the embeddings with the same stack of networks.  The approach has a connection to the kernel-based support vector machine. The well-known Radial Basis Function Kernel (RBF)  essentially is an infinite sum over polynomial kernels, each of which can be further expanded to linear dot products in the polynomial space. The RBF kernel defines a high-dimensional transform $\Psi_{RBF}: K_{RBF} (\textbf{x},\textbf{y}) = \langle \Psi_{RBF}(\textbf{x}), \Psi_{RBF}(\textbf{y}) \rangle$.
An appropriate embedding transformation $\Psi$  approximates  non-linear kernels with linear dot products among embedding vectors, and thus, greatly simplify the downstream relationship learning.  The embedding learning algorithm starts with a random signal and adds incrementally new information into the embedding vectors in multi-layered network architecture in Figure \ref{fig:resnet}.

\subsection{Multi-Layer Model for Tensor Factorization}

\begin{SCfigure*}
\includegraphics[width=0.6\textwidth]{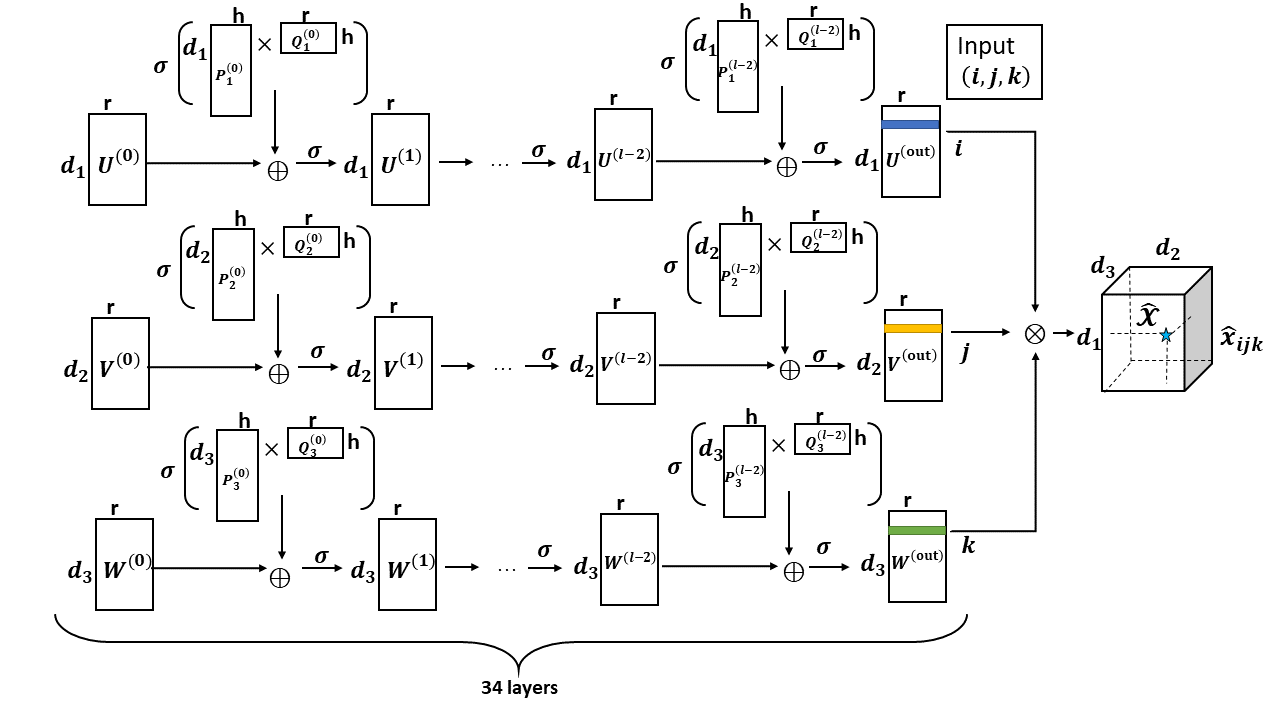}
\caption{In the Multi-Layer Network Architecture for Learning Embedding, we use by-pass connections to create very deep networks (up to 34 layers in this example) to learn complex data structures. The multi-way relationships among all factors are modeled by the linear dot product. We can also use  MLP or convolutional neural networks and trade-off the complexity between the embedding layers and relationship modeling layers. This architecture mitigates the overfitting problem by adding structural constraints in the high-dimensional embedding.}
\label{fig:resnet}
\vspace{-0.1in}
\end{SCfigure*}

We adopt Multi-Layer matrix factorization to construct the factor matrices of the tensor, thereby learning meaningful embeddings. Given a factor matrix $U \in \mathbb{R}^{d_1\times r}$, $U=[\mathbf{u_1}, \mathbf{u_2}, \cdots, \mathbf{u_{d_1}}]^\top$ is a collection of $d_1$ embeddings with $r$ dimensions.  The factor matrix holds the feature vectors of $d_1$ entities. Their feature vectors presumably have structure and are generated from $H$ hidden variables in $l$ different groups (clusters).  For simplicity, we assume that each group uniformly has $h=\frac{H}{l}$ hidden variables.  For example, the real space image features might originate from different groups of hidden variables: frequency bands, pose features, color features, expression features, and identify features.  Based on this preassumption, we further decompose $U$ into two hidden matrices $P$ and $Q$ and learn the feature grouping structure simultaneously as follows:
\begin{equation}
U= P_{d_1 \times H} Q_{H\times r} = \sum_{j=0}^{l-1} P^{(j)} \times Q^{(j)},    \label{eqn:matrix-factorization}
\end{equation}
 where $P^{(j)} =[\mathbf{p}_{jh}, \mathbf{p}_{jh+ 1}, \cdots, \mathbf{p}_{jh + h-1}]^\top$ and  $Q^{(j)} =[\mathbf{q}_{jh}, \mathbf{q}_{jh+ 1}, \cdots, \mathbf{q}_{jh + h-1}]^\top$. When the group information is known, we explicitly arrange the order of hidden variables and provide a group-aware matrix factorization, as shown in the right hand in Eqn \ref{eqn:matrix-factorization}.   In most cases, the group information and hidden variables are unknown, and nevertheless can be extracted by our proposed multi-layer matrix factorization networks.  
 
 The multi-layer matrix factorization has the fundamental connection to signal processing: the $Q^{(j)}$ consists of the base (loading) vectors of transformation (for example, Fourier or Spectral transformation) and $P^{(j)}$ is the loading scores of $U$ on the base matrix $Q^{(j)}$, i.e., the row vectors in $U$ are the weighted sum of the base signals in different frequency bands. We do not assume any prior knowledge of the bands of hidden variables and treat data as the signals from different frequency bands.  Here, each rank of the matrix factorization $P$ and $Q$ represents one frequency band. We will use the $j$-th layer neural network in Figure \ref{fig:resnet}
 to learn the $P^{(j)}$ and $Q^{(j)}$ in Eqn \ref{eqn:matrix-factorization} and attempt to extract $h$ related frequencies in the same band simultaneously.   The dimensonality of $U$ at different layer is the same ($U^j  \in \mathbb{R}^{d_i\times r}$ and $U^{j-1} \in \mathbb{R}^{d_i\times r}$), which helps to remove noises and learn meaningful signal without forcing the model to compress the available information.  
 
Given the complicated relationship embedded in $U$,  this multi-layer approach partitions the learning into $l$ frequency bands. Each frequency band corresponds to a network layer in Figure \ref{fig:resnet}.  This design eases the complexity associated with single layer and avoids the complexity of learning to model the entire signal altogether.


\subsection{Non Linearity and By-pass Connection}
We introduce non-linear transfer functions $\sigma$, i.e., ReLU, ELU, and sigmoid, in the factor matrix $U$ and its corresponding multi-layer neural network of Eqn \ref{eqn:matrix-factorization}. 
We rewrite non-linear matrix  factorization in each layer $j$  as follows:

\begin{equation}
U^{(j)} = \sigma (U^{(j-1)} + \sigma (P^{(j)} \times Q^{(j)})),  
\label{eqn:non-linear-matrix-factorization}
\end{equation}
where the input matrix at layer $0$ is zero, and the inputs to the multi-linear dot product in the right-hand side of Figure \ref{fig:resnet} is $U^{(out)}=U^{(l-1)}$.  Figure \ref{fig:resnet} shows $P^{(j)}$ and $Q^{(j)}$ are the trainable parameters, and their products are applied with non-linear transfer functions before and after being added into the forward path from the lower layer to the higher layer.  The network has a by-pass connection from the layer input directly to the layer output and applies element-wise matrix additions to implement identity mapping. Similar to the ResNet \cite{he2016deep}, the by-pass connection design does not increase the number of neurons and mitigates the problem of gradient vanishing and explosions that often occur in networks with a great depth. The multiple layers of matrix factorization and by-pass greatly enhance the modeling capacity for non-linearity, while incurring no higher training errors than those without it.  

\IncMargin{0.1em}
\begin{algorithm2e}[t]

\newcommand\mycommfont[1]{\footnotesize\ttfamily\textcolor{black}{#1}}
\SetCommentSty{mycommfont}
\SetKwData{Y}{Y}
\SetKwFunction{Fold}{Fold}\SetKwFunction{Loss}{Loss}
\SetKwInOut{Input}{Input}\SetKwInOut{Output}{Output}
\Input{Tensor $\mathcal{X} \in \mathbb{R}^{d_1 \times  d_2  \times d_3}$ and tensor $\mathcal{Y} \in \mathbb{R}^{d_1 \times  d_2  \times d_4}$ to be completed,
rank of tensor decomposition $r$, rank of matrix factorization $h$, network layers $l$, index set of observed entries $\Omega_{\mathcal{X}}$ in the tensor $\mathcal{X}$ and $\Omega_{\mathcal{Y}}$ in the tensor $\mathcal{Y}$.\\
}
\Output{Updated factor matrices $U^{(0)}, V^{(0)}, W^{(0)}, T^{(0)}$ and hidden matrices $P^{(j)}$ and $Q^{(j)}$ $(j = 0, \dots,l-1)$}
\BlankLine
\emph{Initialize all hidden matrices $P^{(j)}$ and $Q^{(j)}$  for all layers, initialize $U^{(0)}, V^{(0)}, W^{(0)}, T^{(0)}$}\\
\Repeat{maximum number of epochs or early stopping}{
\For{$\alpha = \forall(i,j,k) \in \Omega_{\mathcal{X}} \cup \Omega_{\mathcal{Y}}$}{
\tcp{Forward Propagation} 
\For{$j\leftarrow 1$ \KwTo $l-1$}{
$U^{(j)} = \sigma (U^{(j-1)} + \sigma (P^{(j)} \times Q^{(j)}))$ \\
Similarly, calculate $V, W, T$ \tcp*{Eqn \ref{eqn:non-linear-matrix-factorization}}
}
Calculate loss $\mathcal{L}$ \tcp*{Eqn \ref{eqn:SGD-CMTF}} 
\tcp{Backward Propagation}  
\eIf{$\alpha \in \Omega_{\mathcal{X}}$}{
    Update $U^{(0)}, V^{(0)}, T^{(0)}$
    and associated $P^{(j)}$ and $Q^{(j)}$ 
    based on chain rule and Eqn. \ref{eqn:gradient_X}.
   }{ Update all $U^{(0)}, V^{(0)}, W^{(0)}$ and
   associated $P^{(j)}$ and $Q^{(j)}$ based on chain rule and Eqn. \ref{eqn:gradient_X}.}
}}
\caption{MLCTR Coupled}\label{MLCTR_Algorithm}
\end{algorithm2e}\DecMargin{1em}
\vspace{-0.1in}
 
\subsection{Coupled Tensor Factorization for Embedding Learning and Prediction}
Nearly all tensor completion algorithms, including our proposed MLCTR, often suffer the cold start problem and extremely low signal to noise ratio (SNR) \cite{acar2012coupled}.  Especially in our finance application, the EPS dataset has a high number of missing values, i.e., 99\%. 
The $time$, $analyst$, and $firm$ latent factors learned from the EPS tensor are less informative because of the excessive number of miss values. To recover the critical missing signal, we introduce additional data synergistic to the EPS tensor to be imputed. The firm fundamentals share the same time and firm dimensions with EPS and provide complementary information for any firm in EPS, including key performance indicators and firm characteristics. 

\begin{figure}[!t]
\centerline{\includegraphics[scale=0.28]{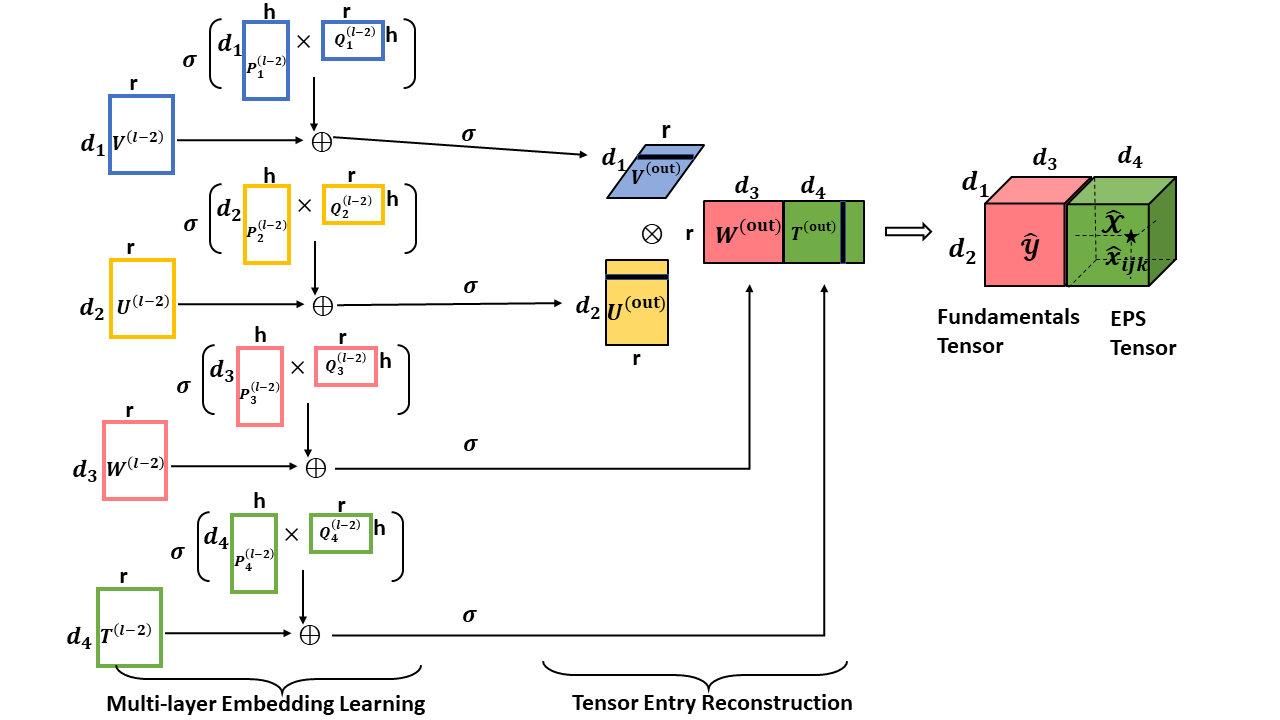}}
\caption{Coupled Tensor Factorization and Completion}
\label{coupledtensor}
\vspace{-0.2in}
\end{figure}

In the coupled tensor framework, MLCTR attempts to enforce the same time factor and firm factor matrices during factorization.  The information propagates from the dense tensor (firm accounting fundamentals) to the sparse tensor (analysts' EPS forecast) by coupling the firm and time factors in tensor factorization and completion.  Figure \ref{coupledtensor} describes the MLCTR system for coupled tensor, factorizing two tensors: $\mathcal{X}$ and $\mathcal{Y}$, each of which has three factor matrices: two common factor matrices $U$ and $V$ and the unique matrix $T$ for $\mathcal{X}$ and $W$ for $\mathcal{Y}$.  We assume all factor matrices have the same rank $r$. 
Figure \ref{coupledtensor} shows a simple linear dot product for multi-way relationship. Alternatively, we add MLP between the embedding learning layer and the output layer for modeling addition non-linear relationships.\footnote{In the experiment part, we call MLCTR for using simple dot product and MLCTR (MLP) for using MLP layers between the middle layer and output layer on the architecture.} 
The objective function for coupled tensor is to minimize the mean square error of two tensor factorizations by optimizing the following equation: 
\begin{align}
 \label{eqn:CMTF}
    \mathcal{L}= \big|\!\big|\mathcal{X}- [\![\Lambda_1; U^{(out)}, V^{(out)}, T^{(out)}]\!] \big| \!\big|_F^2 
    \\
\notag    + \lambda \big|\!\big|\mathcal{Y}   - [\![\Lambda_2; U^{(out)}, V^{(out)}, W^{(out)}]\!] \big| \!\big|_F^2
\end{align} 
Here $\lambda$ is the hyper-parameter to adjust the relative importance between the two coupled tensors.  The factors matrices $T^{(out)}, U^{(out)}, V^{(out)}, W^{(out)}$ are the output of the multi-layer embedding learning networks. 

\subsection{SGD-based Coupled Tensor Factorization and Integration} 
Traditional CMFT requires complete tensors for factorization and incurs long computation time for large tensors. In this paper, we perform an element-wise tensor reconstruction from the observation set $\Omega$ for fast convergence.  We revise the objective function as follows to be compatible with any deep neural network platform:


\vspace{-0.15in}
\begin{align}
\label{eqn:SGD-CMTF}
    \mathcal{L} =  \sum_{i,j,k \in \Omega_{\mathcal{X}} \cup \Omega_{\mathcal{Y}}}
        \mathbbm{1}_{\Omega_{\mathcal{X}}} (ijk) \big|\!\big|\mathcal{X}_{ijk}-  \sum_{s=1}^r U_{is}^{(out)} V_{js}^{(out)} T_{ks}^{(out)} \big| \!\big|_F^2  \\
 \notag        + \lambda \mathbbm{1}_{\Omega_{\mathcal{Y}}} (ijk) \big|\!\big|\mathcal{Y}_{ijk} - \sum_{s=1}^r  U_{is}^{(out)} V_{js}^{(out)} W_{ks}^{(out)}) \big| \!\big|_F^2,
\end{align}
where $\mathbbm{1}_{\Omega_{\mathcal{X}}} (ijk)$ is an indicator function. It is straightforward to implement Eqn \ref{eqn:SGD-CMTF} with the Stochastic Gradient Descent (SGD): we first mix the training data from $\Omega_\mathcal{X}$ and $\Omega_\mathcal{Y}$, randomly choose one mini-batch of training samples, and calculate the mean square error defined in Eqn \ref{eqn:SGD-CMTF}. 
This mixing strategy intelligently employs indicator functions to allow any observation to be treated uniformly, thereby enabling the parallel processing of the samples in the same mini-batch. Eqn \ref{eqn:SGD-CMTF} essentially is multi-task learning and thereby highly scalable to allow multiple tensors to be factorized simultaneously.

For any training sample, the gradient of the first term or the second term in Eqn. \ref{eqn:SGD-CMTF} is zero.  
Considering the sample with index $(i,j,k) \in \Omega_\mathcal{X}$, we calculate the corresponding gradient of $\mathcal{L}$ to the embeddings and update the relevant parameters  as follows:

\begin{equation}
\begin{aligned}
\label{eqn:gradient_X}
U_{i,:}^{(out)}& \leftarrow U_{i,:}^{(out)}\\
 & - \eta ( \sum_{s=1}^r U_{is}^{(out)} V_{js}^{(out)} T_{ks}^{(out)}- \mathcal{X}_{ijk}) (V_{j,:}^{(out)} \odot T_{k,:}^{(out)})
\end{aligned}
\end{equation}
Here $\odot$ is the Hadamard product of two vectors. The gradients on other factor matrices $V,W,T$ have the identical formula to Eqn \ref{eqn:gradient_X}. We only show the parameter in matrix $U$. The gradient is back-propagated through the network layers in Figures \ref{fig:resnet} and \ref{coupledtensor}. Algorithm \ref{MLCTR_Algorithm} shows the pseudo-code of SGD based MLCTR\footnote{We will publish the Python notebooks in GitHub for reproducibility.}.

\section{Experiments}
We conduct two experiments for four datasets to evaluate our algorithm: 1) efficiency in tensor completion in both time and accuracy compared to other state-of-the-art tensor completion techniques and 2) ability to factorize sparse coupled tensor while learning meaningful factor matrix. To alleviate the overfitting problem, we try several regularization methods inlcudung Lasso, Ringe, and Elaticnet and call these models as Resnet-L1, Resnet-L2, Resnet-Elastic. We compare our algorithm with CPWOPT \cite{acar2011scalable} - the benchmark low-rank sparse tensor completion method, P-Tucker \cite{oh2018scalable} - a scalable Tucker model with fully paralleled row-wise updating rule, and CoSTCo \cite{liu2019} - CNN based state-of-the-art nonlinear tensor completion method. To evaluate performance, we use three metrics, RMSE, MAE, and MAPE.

\subsection{Data}
We apply our MLCTR for SafeGraph Foot Traffic data. SafeGraph collects cellphone GPS location data from a panel of cellphone 
users when a set of installed apps are used and they are available for free to academics studying COVID-19 (https://www.safegraph.com/covid-19-data-consortium). These cellphone GPS location data 
are supplied at the daily level for residents of each Census Block Group(CBG). In the following experiment We collect 95509754 records belonging to five boroughs of New York States (The Bronx, Brooklyn, Manhattan, Queens, and Staten Island.) for the sample period of 2019. And we use these records to construct a three order tensor to describe the three-way relationship of original CBG, destination CBG and date with shape $6439 \times 6439 \times 365$. To ensure a reasonable data distribution, we apply the log transformation and use the grid search to find the proper base as 10.

\begin{table}

\caption{Data statistics and Hyper-parameters }
\vspace{-0.2in}
\begin{center}
\resizebox{\linewidth}{!}{%
\begin{tabular}{ l c c c c c}
\toprule

Datasets & shape & observed entries & lr &  batch size  \\
\midrule
SafeGraph  &  (6439, 6439, 365) & 95509754 & 1e-4 & 8192 \\
SafeGraph (log10)  &  (6439, 6439, 365) & 95509754 & 1e-4 & 8192 \\

\bottomrule
\end{tabular}}
\label{tab:Data}
 \vspace{-0.2in}
\end{center}
\end{table}

In addition, we test the efficiency of our algorithm on two commonly used public datasets. The first one is climate data, which is used in \cite{lozano2009spatial, liu2010learning}. The dataset has 18 climate agents from 125 locations from 1992-2002. The second one is a real point of interest (POI) data used in \cite{li2015rank}. The Foursquare check-in data made in Singapore between Aug. 2010 and Jul. 2011. The data comprises 194,108 check-ins made by 2,321 users at 5,596 POI. Using two different processing systems, we develop two different tensor representations of the POI dataset, i.e., POI and POI-3D. For POI we follow the approach used in \cite{liu2019} and represent the tensor as $(user\_id, poi\_id, location\_id )$. The first two dimensions $user\_id$ and $ poi\_id $ are available in data, in \cite{liu2019}, the authors created the third dimension `$location\_id$' by splitting the POIs into 1600 location clusters based on their respective latitude and longitude. Hence, both 2nd-order and 3rd-order represent location information, and for each $location\_id$, we have different $poi\_id$, resulting in an unnecessary large tensor. In POI-3D we overcome this limitation by incorporating time information available in the data and replace $location\_id$ with $time$ to represent tensor as $(user\_id, poi\_id, time)$.
We divide the 24 hours into 12 groups of 2 hours intervals. This incorporation of time information helps us learn a better latent representation of user probability of visiting a specific POI at a specific time. 

We normalize the datasets with zero mean and unit variance. For EPS and climate data, we use an 80/20 train-test split, with 10\% of the training data as the validation set and early-stopping if validation loss does not improve for 10 epoch. For both POI datasets, we use the train-validation-test set following \cite{li2015rank}. The tensor shape, number of observed entries for each of the data sets, and Hyper-parameters are reported in Table \ref{tab:Data}.

\begin{table*}
\caption{ Tensor Completion Result }
\vspace{-0.2in}
\begin{center}
\resizebox{\linewidth}{!}{%
\begin{tabular}{l | l | c c c c | c c c c| c c c c }
\toprule
 &  \multicolumn{1}{l|}{Metric}  &  \multicolumn{4}{c|}{RMSE} &  \multicolumn{4}{c|}{MAE} & \multicolumn{4}{c}{MAPE}
 \\
\midrule
Data & Model/rank & 10 & 20 & 30 & 40 & 10 & 20 & 30 & 40 & 10 & 20 & 30 & 40

\\
\midrule 
\multirow{4}{*}{\shortstack[l] {EPS and \\ Fundamentals }}  & CPWOPT	& 0.3865 &  0.3434 &	0.3352 &	0.4171 &	0.1735 &	0.1722 &	0.2274  &	0.2137 &	35.7778  &	36.2522 &	32.3325 &	39.8641

\\

& P-Tucker & 0.3364 &	0.3106 &	0.2954 &	0.2802 &	0.2173 &	0.1934 &	0.1755 &	0.1406 &	34.7840 &	32.1949 &	29.1234 &	28.0701
\\

& CoSTCo & 0.2536 &	0.2364 &	0.2455 &	0.2310 &	\textbf{0.1337} &	0.1338 &	0.1185 &	0.1096 &	29.7298 &	30.4989 &	26.4746 &	24.0901
\\
& MLCTR &	0.3229 &	0.2900 &	0.1947 &	0.1818
&	0.1678 &	0.1457  &	0.0984  &   0.0946 &	31.4561
&	28.6817  &	21.7898 &	21.1656
\\

& MLCTR (MLP) & 0.2645 &	0.2140  &	0.1940 &	0.1996 &	0.1385 &	0.1078  &	0.0979 &	0.0914 &	27.6427 &	22.2719 &	21.1212  &	19.4139

\\
& MLCTR (Coupled) & 	\textbf{0.2421}&	\textbf{0.2065} &	\textbf{0.1703} &	\textbf{0.1520 }&	0.1355 &	\textbf{0.1016 }&	\textbf{0.0810} &	\textbf{0.0806} &	\textbf{24.2524} &	\textbf{18.9695} &	\textbf{16.6764 }&	\textbf{17.4226 }
\\
\midrule 

\multirow{3}{*}{Climate}   
& CPWOPT & 	0.4155 &	0.3636  &	0.3363 & 0.3294  &	0.2694 &	0.2144 &	0.1869 &	0.1754 &	94.4968 &	94.3807 &	87.7374  &	86.0715 \\

& P-Tucker &  	0.4231 &	0.3746  &	0.3224 & 0.3201  &	0.2844 &	0.2417 &	0.1942 &	0.1648 &	104.4845 &	98.7135 &	78.4751  &	75.5421

\\

& CoSTCo & 	0.3955  &	0.3040 &	0.3019  &	0.2577 &	0.2676  &	0.2022 &	0.1991 &	0.1667 &	59.7978 &	47.7849&	47.4389 &	39.6156

\\ & MLCTR &  0.3750 & 0.3023 & \textbf{	0.2543 }& 	\textbf{0.2326 }& 	0.2501	& 0.1953 & \textbf{	0.1625 }& \textbf{0.1434 }&  55.7723 & \textbf{44.1495 }&  \textbf{	37.9093} & 	\textbf{33.3923}

\\
& MLCTR (MLP) & 	\textbf{0.3558} & \textbf{0.2902} &	0.2572 &	0.2410 & \textbf{0.2378} & \textbf{0.1897} & 0.1679 &	0.1551 &	\textbf{54.2780} & 45.0224	& 40.5167 &	37.5005
\\

\midrule 
\multirow{3}{*}{POI} &  P-Tucker & 0.2464 & 	0.2182 & 	0.1989 &  0.1784 & 	0.1484 &	0.1443 & 	0.1204 & 0.1182 &  84.1285 &	81.4561 &	72.9842 & 	70.9242

\\
& CoSTCo & 	0.1536 & 	\textbf{0.1532 }& 	0.1535 &  0.1534 & 	0.0887 &	0.0877 & 	0.0883 & 0.0849 &  54.8782 &	53.5907 &	54.4279 & 	49.9463 
\\
& MLCTR & \textbf{ 0.1532 }&	0.1534 &	\textbf{0.1531}  & 	\textbf{0.1533}	& 0.0883 & \textbf{0.0819} & 0.0878 & 	0.0846	 & 54.5828 & \textbf{47.1175} &  54.1197	& 50.0688
\\

& MLCTR (MLP) &	0.1539 &	0.1538 & 0.1536  &	0.1541 &	\textbf{0.0833} &	0.0829 &	\textbf{0.0829} &	\textbf{0.0826} & \textbf{48.1277} &	47.5316 &	\textbf{47.5259} &	\textbf{46.8857}\\ 
\midrule

\multirow{3}{*}{POI-3D}  &  P-Tucker & 	0.1814 &	0.1784 &	0.1441 &	0.1403 &	0.1103  &	0.1017 &	0.0987 &	0.0915 &	57.0812 &	46.9813 &	42.7891 &	39.8714

\\
& CoSTCo  &	0.1064 &	0.1056 &	\textbf{0.1057} &	0.1052 &	0.0543  &	0.0517 &	\textbf{0.0521} &	0.0497 &	36.0018 &	33.5342 &	\textbf{33.9658 }&	31.1371

\\

& MLCTR & 	0.1084 &	0.1088 &	0.1064  &	0.1064 &	0.0564 &	0.0566 &	0.0559 &	0.0571 &	37.1333 &	37.1333 &	37.2212  &	38.3884  
\\

& MLCTR (MLP) &	\textbf{0.1051} &	\textbf{0.1053 }&	0.1058 &	\textbf{0.1052} &	\textbf{0.0486} & \textbf{0.0504} &	0.0535&	\textbf{0.0495} &	\textbf{30.2706} & \textbf{32.1058 }&	35.3082 &	\textbf{31.0428} \\
\bottomrule
\end{tabular}
}
 \vspace{-0.2in}
\label{tab:Result_rank}
\end{center}
\end{table*}

\subsection {Coupled Tensor Completion}
\begin{figure}
  \centering
  \vspace{-0.1in}
  \subfloat[Convergence Plot\label{fig:Convergence}]{\includegraphics[width=.50\linewidth]{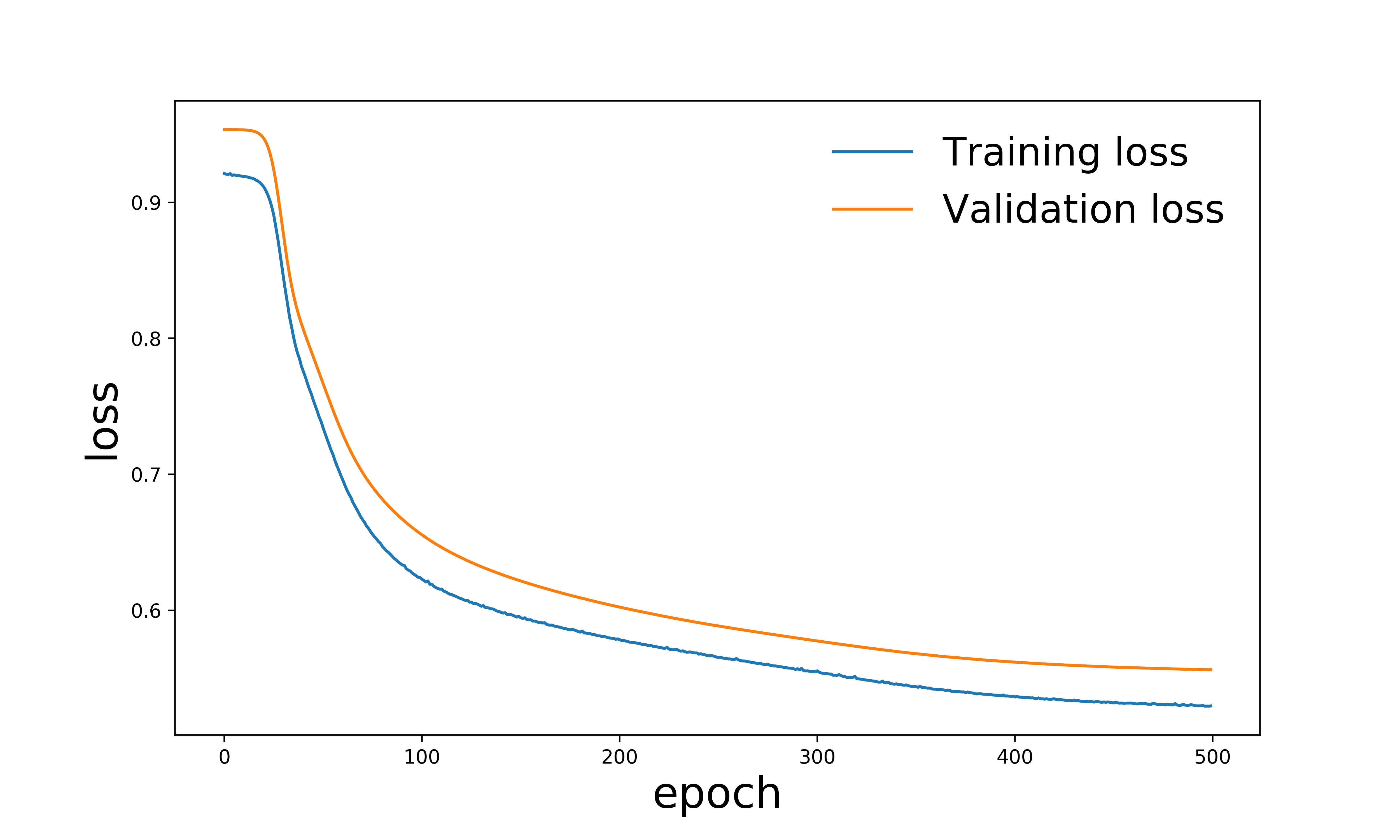}}
  \subfloat[Rank vs Accuracy\label{fig:rank_fund}]{\includegraphics[width=.50\linewidth]{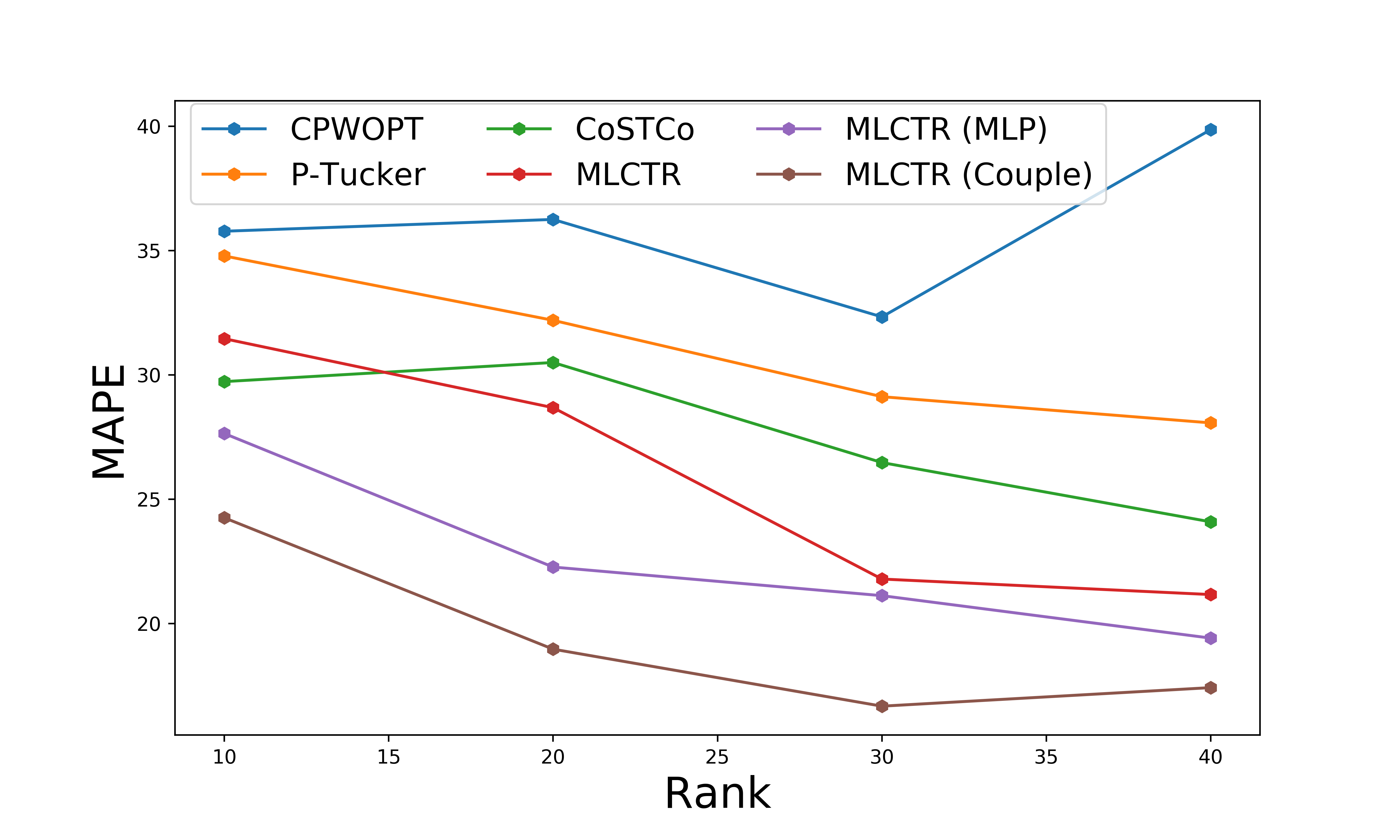}}
  
   \caption{ (a) Convergence plot of MLCTR (Coupled) for EPS data. (b) Tensor completion accuracy at different ranks for EPS data.
   }
  \label{fig:Conv_Rank}
   \vspace{-0.2in}
\end{figure}

We factorize two sparse tensors --analysts' EPS forecast $(quarter, firm, analyst)$ and firm fundamentals $(quarter, firm, fundamental)$-- together with the objective function of eq. \ref{eqn:SGD-CMTF}. For imputing missing values, coupled tensor factorization produces much higher accuracy than single factorization. As reported in Table \ref{tab:Result_rank}, MLCTR (coupled) can outperform CPWOPT by 49\% (rank = 30, best performing CPWOPT) and CoSTCo by 34\%(rank = 40, best performing CoSTCo). The benefits of coupled tensor completion beyond single tensor completion can be captured by the performance improvement between MLCTR and MLCTR coupled. With rank 40, MLCTR (coupled) outperforms MLCTR (MLP) by 16\% (RMSE), 15\% (MAE) and 18\% (MAPE). The proposed MLCTR algorithm is also robust to the increasing number of missing values. As shown in Figure \ref{fig:tens_com}, even with 99\% missing values, our algorithm can still impute missing values accurately, outperforming CoSTCo by 37\%. MLCTR is also less sensitive to rank. Figure \ref{fig:rank_fund} shows that with higher ranks, MAPE decline smoothly for all three versions of MLCTR. 

\subsection {Sparse Single Tensor Completion}
MLCTR is not only effective for coupled tensor completion, but the technique of using residuals by further factorizing latent factors can also learn better embedding for single tensor. To show such generalization of MLCTR, we conduct analysis using three public datasets. 
On climate forecasting, MLCTR outperforms CPWOPT, P-Tucker, and CoSTCo in all three performance metrics (Table \ref{tab:Result_rank}).  At rank 30, MLCTR (MLP) outperforms CPWOPT by 24\%, P-Tucker by 20\% and CoSTCo by 15\% in RMSE. 

For both POI datasets, the data sparsity is too high. With only 0.0005\% (POI) and  0.09\% (POI-3D) available observation, CPWOPT with gradient descent does not converge. Therefore, for POI data, we did not report the CPWOPT result. In POI, with rank 30, MLCTR (MLP) outperforms P-Tucker by 31\%, and CoSTCo by 13\% in MAPE; whereas CoSTCo is only better at rank 10 in RMSE.  For POI-3D, CoSTCo outperforms simple MLCTR in some performance metrics. However, our MLP version MLCTR (MLP) still outperforms CoSTCo with higher ranks (30 and 40) by a significant margin.

\begin{figure}[!t]
\begin{centering}
  \includegraphics[width=.65\linewidth]{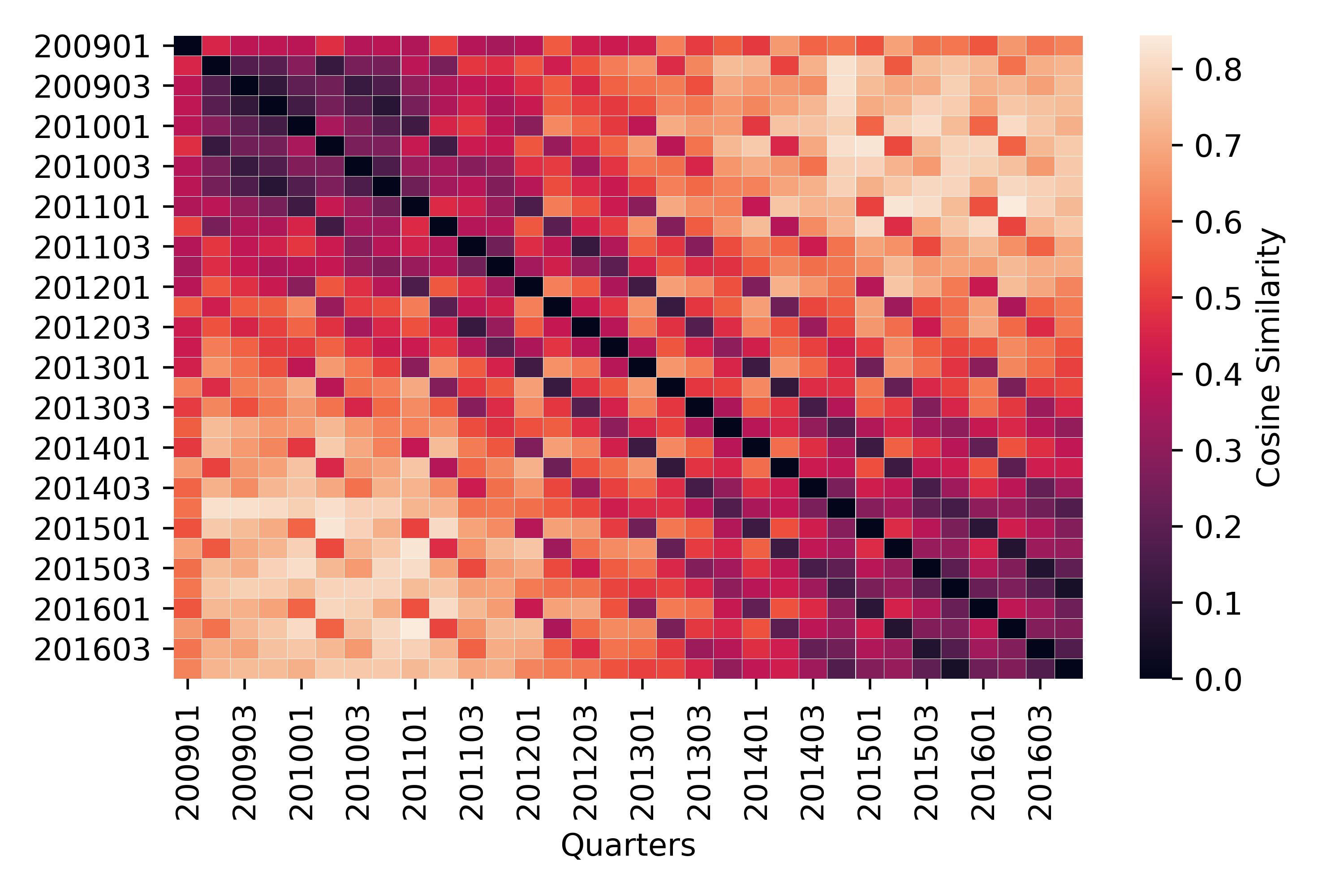}
  \vspace{-0.1in}
    \caption{Similarity heatmap among ``quarters” factors.}
  \label{fig:heatmap}
  \vspace{-0.1in}
 \end{centering}
\end{figure}

\begin{figure}[!t]
\begin{centering}
  \includegraphics[width=.90\linewidth]{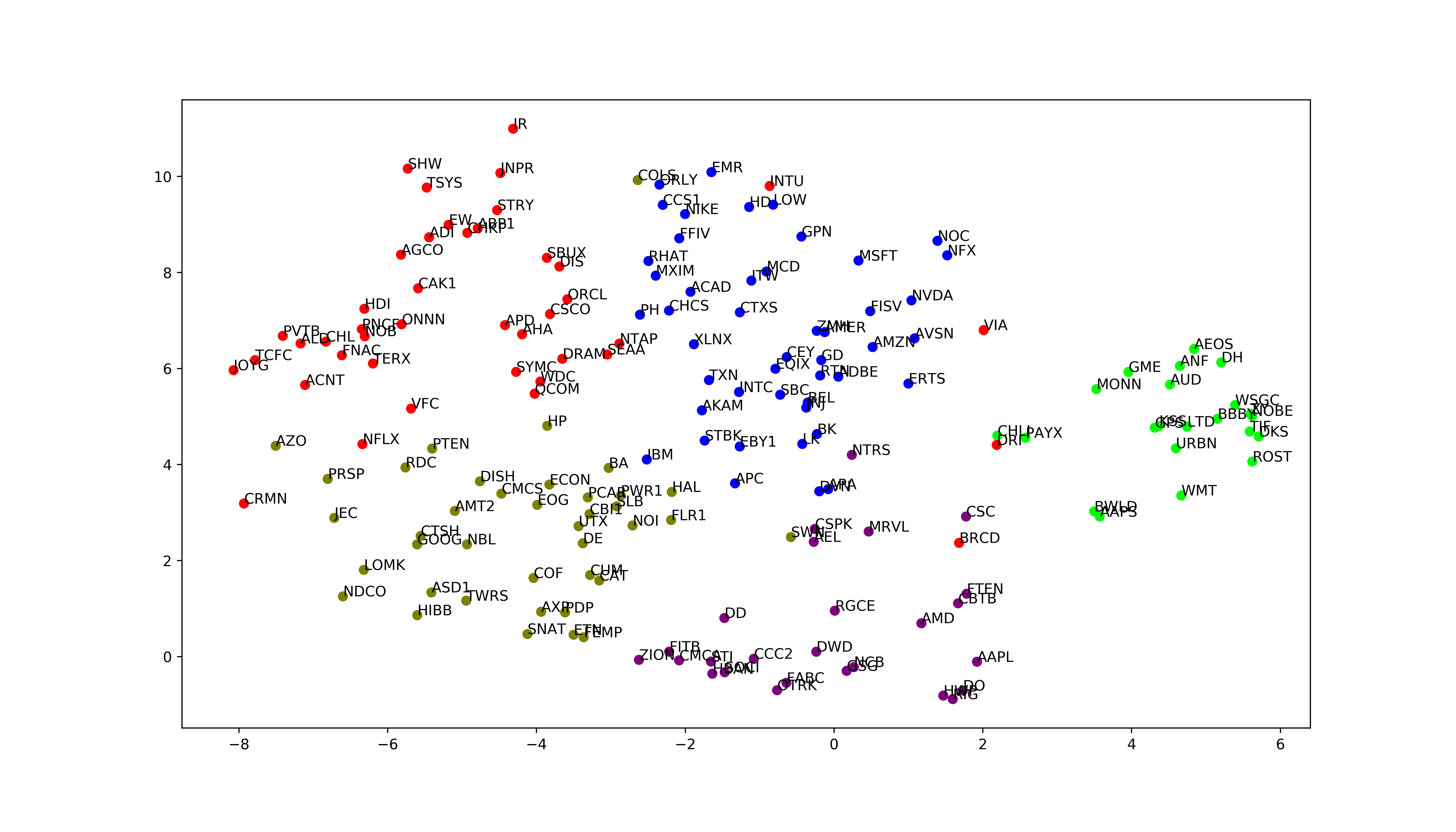}
  \vspace{-0.2in}
    \caption{t-SNE and spectral clustering on learned ``firm" factors.}
  \label{fig:similarity}
  \vspace{-0.1in}
 \end{centering}
\end{figure}

\subsection{Visualization of Learned Factors}
As shown in Fig. \ref{fig:embedding}, the learned factor matrices from MLTCR is much more informative than other nonlinear tensor factorization models. To further understand the learned factors, we visualize the factor matrices learned from coupled tensor factorization on analysts' EPS forecast and firm fundamentals data. Fig. \ref{fig:heatmap} shows the cosine similarity between ``quarters" learn from the time factors matrix. The temporal patterns in the time factors are clearly visible. Fig. \ref{fig:factor_vis} shows the t-distributed stochastic neighbor embedding (t-SNE) to plot the spectral clustering based on firm latent factors. MLCTR learns meaningful embedding for firms according to their size, service type,  and the client groups they serve. For example, major retail brands like Walmart (WMT), Urban Outfitters (URBN), Gap (GPS), Abercrombie \& Fitch (ANF) are grouped Far-right (Green); major tech companies like Microsoft (MSFT), NVIDIA (NVDA), Amazon (AMZN), IBM, Akamai (AKAM) are grouped in top-center (Blue); and financial service companies like Fidelity (FABC), Zions Bancorporation (ZION), Morgan Stanley (DWD), iShares (GSG) are grouped in bottom-center (Purple).

\begin{figure}[!t]
\begin{centering}
  \includegraphics[width=1\linewidth]{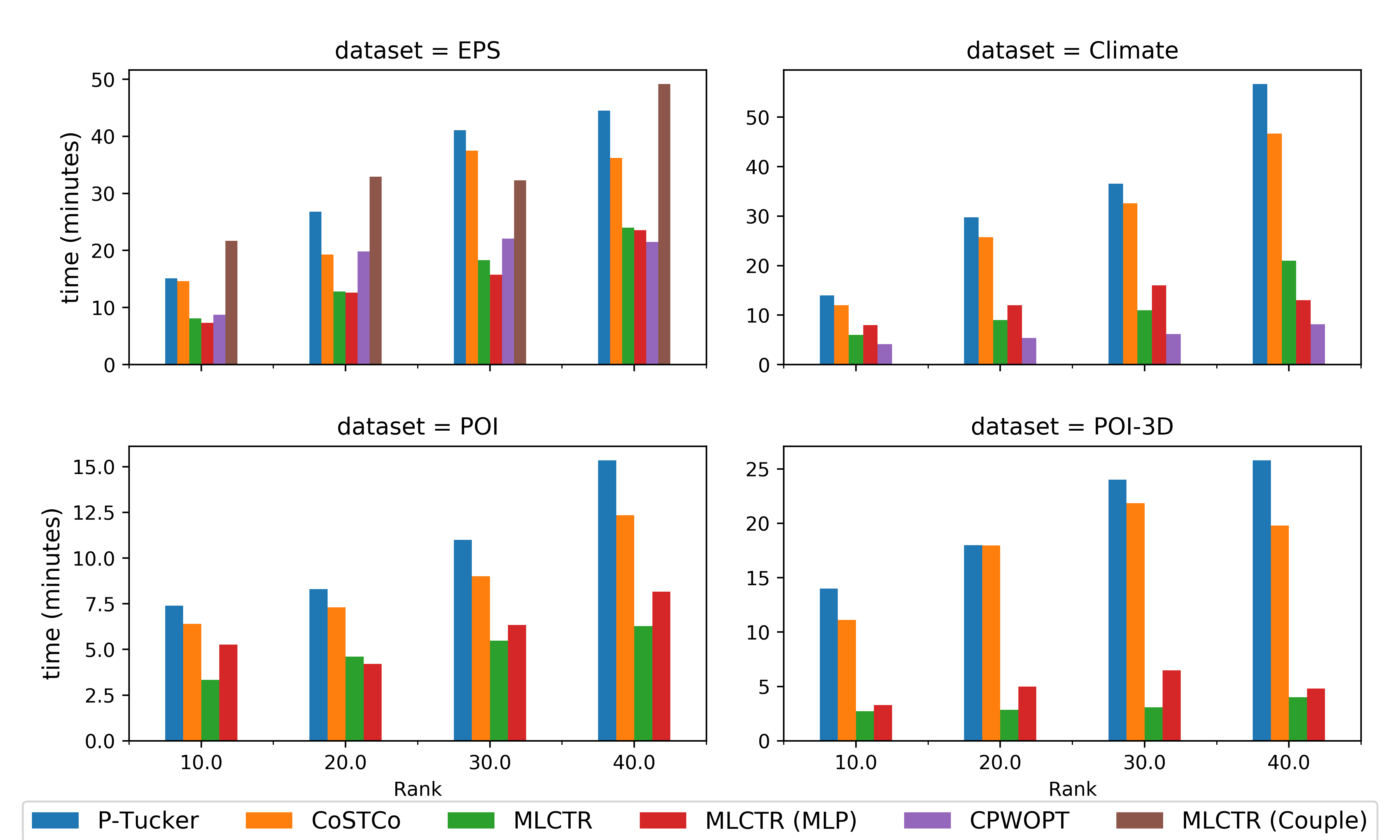}
    \caption{Running time of different tensor completion algorithms at different ranks.}
  \label{fig:time}
  \vspace{-0.1in}
 \end{centering}
\end{figure}

\subsection{Running Time Comparison}
MLCTR uses low-rank MF and by-pass connections for learning latent factors; thus, it learns embedding matrix much faster than other nonlinear algorithms, i.e., CoSTCo, P-Tucker. MLCTR takes indices of the observed values as input variables; therefore, it is linear to the number of available observations rather than the size of the target tensor. 
Figure \ref{fig:time} shows the running time of each algorithm in each data sets at different ranks. The reported time elapsed for each algorithm is with early stopping criteria. The time complexity of MLCTR is also linear to the rank and does not increase drastically with higher ranks. 

\section{Discussion}
The proposed MLCTR algorithm divides the tensor factorization and completion into two interleaved modules: the first one that learns the rank $r$ embeddings and the second one for modeling multi-way relationships among the embeddings of the participating entities.  The majority of related work focuses on the latter: for an $N$th-order tensor, $N$-way linear (including CP and Tucker decompositions) and nonlinear kernels (RBF, polynomial) are employed to model the relationships and minimize the mean square errors between the observations and predicted values.   

Tensor rank is the key parameter in factorizing a tensor. The common practice is to perform a grid search on an appropriate rank $r$. A small rank $r$ incurs large bias in tensor analysis while a high rank $r$ leads to overfitting \cite{liu2019}.  The overfitting problem is primarily due to many unconstrained $r$ hidden variables in embeddings and must be regularized to minimize variance.   The standard $l_1$ and $l_2$ regularizations only add the local constraints of smoothness and sparsity on embeddings and might not be sufficient for our problem.  Inspired by the signal processing theory, we introduce the structure, base constraints, and global regularization to the embedding space. We argue that high-quality embedding learning will mitigate the complexity in the second module for relationship modeling so that a simple linear dot product in CP or a shallow MLP is sufficient in the algorithmic implementation.

\section {Conclusion}
In this paper, we apply an innovative approach to shift the learning towards the embedding module, ensuring its central role in a tensor algorithm, and easing relationship learning.  With the high-quality embedding, many multi-way relationships can be efficiently modeled by the CP tensor algorithm or simple MLP networks. We implement MLCTR using multi-layer neural networks where each layer performs low-rank matrix factorization for embedding matrices. Experiments show that our algorithm works exceptionally well for both single tensor and coupled tensor factorization and completion and is less sensitive to tensor rank, robust to noise, and fast to converge during training.

\bibliographystyle{apalike}

\bibliography{ref}

\end{document}